\documentclass{article}
\usepackage{spconf}
\usepackage{amsmath,amssymb,mathtools,bm}
\usepackage{graphicx}
\usepackage{booktabs}
\usepackage{multirow}
\usepackage{microtype}
\usepackage{siunitx}
\usepackage{enumitem}
\usepackage{xcolor}
\usepackage{url}
\usepackage{placeins}

\def\Title{A Comparative Study of Machine Learning and Deep Learning for Out-of-Distribution Detection}
\def\Authors{Jihyeon Baek \quad Seunghoon Lee \quad Gitaek Kwon \quad Doohyun Park$^{\dagger}$%
\thanks{$^{\dagger}$Corresponding author: \texttt{doohyun.park@vuno.co}}%
\thanks{Accepted at IEEE International Symposium on Biomedical Imaging (ISBI) 2026.}}
\def\AffilA{VUNO Inc.}

\title{\Title}
\name{\Authors}
\address{\AffilA}

\begin{document}
\ninept
\maketitle

\begin{abstract}
Out-of-distribution (OOD) detection is essential for building reliable AI systems, as models that produce outputs for invalid inputs cannot be trusted. Although deep learning (DL) is often assumed to outperform traditional machine learning (ML), medical imaging data are typically acquired under standardized protocols, leading to relatively constrained image variability in OOD detection tasks. This motivates a direct comparison between ML and DL approaches in this setting. The two approaches are evaluated on open datasets comprising over 60{,}000 fundus and non-fundus images across multiple resolutions. Both approaches achieved an AUROC of 1.000 and accuracies between 0.999 and 1.000 on internal and external validation sets, showing comparable detection performance. The ML approach, however, exhibited substantially lower end-to-end latency while maintaining equivalent accuracy, indicating greater computational efficiency. These results suggest that for OOD detection tasks of limited visual complexity, lightweight ML approaches can achieve DL-level performance with significantly reduced computational cost, supporting practical real-world deployment.

\end{abstract}


\section{Introduction}
\label{sec:intro}

Real-world computer-aided detection (CADe) and diagnosis (CADx) systems are frequently exposed to out-of-distribution (OOD) inputs that are irrelevant to their intended tasks. OOD detection mitigates this risk by filtering such inputs before downstream inference. In ophthalmology, for instance, non-fundus images—such as external-eye photographs, slit-lamp views, or even chest radiographs—can inadvertently enter fundus imaging workflows. Without proper screening, conventional CADe/CADx systems may still generate outputs for these invalid inputs, producing spurious probabilities and potentially disrupting clinical workflows. This behavior undermines system reliability and highlights the need for explicit OOD detection in safety-critical medical artificial intelligence (AI) applications. While academic benchmarks typically assume fixed in-distribution datasets, real-world deployment requires robustness to unforeseen inputs through integrated OOD detection mechanisms.

The research community has explored a wide range of OOD detection strategies, including confidence thresholds, energy scores, distance metrics in latent space, generative reconstruction errors, and ensemble-based uncertainty estimation \cite{liu2020energy, lee2018simple}. Although these methods have advanced the field, most rely on high-capacity deep encoders and substantial computational resources to achieve adequate performance. Consequently, much of the OOD literature implicitly assumes that the task itself is inherently complex and requires deep learning (DL)–level representation power \cite{yang2024generalized}.

In medical imaging, however, data are generally acquired under standardized conditions with dedicated imaging devices, leading to limited variation in viewpoint, illumination, and anatomical context. As a result, OOD detection between fundus and non-fundus images in this domain may be relatively straightforward. In such settings, a classical machine learning (ML) approach may be both sufficient and preferable, providing comparable accuracy with significantly higher computational efficiency and interpretability.

Based on these assumptions, this study investigates whether the complexity of DL is justified for OOD detection in medical imaging. Using diverse open datasets covering multiple medical imaging modalities and resolutions, we evaluate accuracy, computational efficiency, and robustness, with the aim of assessing the trade-offs between model complexity and practical deployment.

\begin{figure*}[t]
  \centering
  \includegraphics[width=\textwidth]{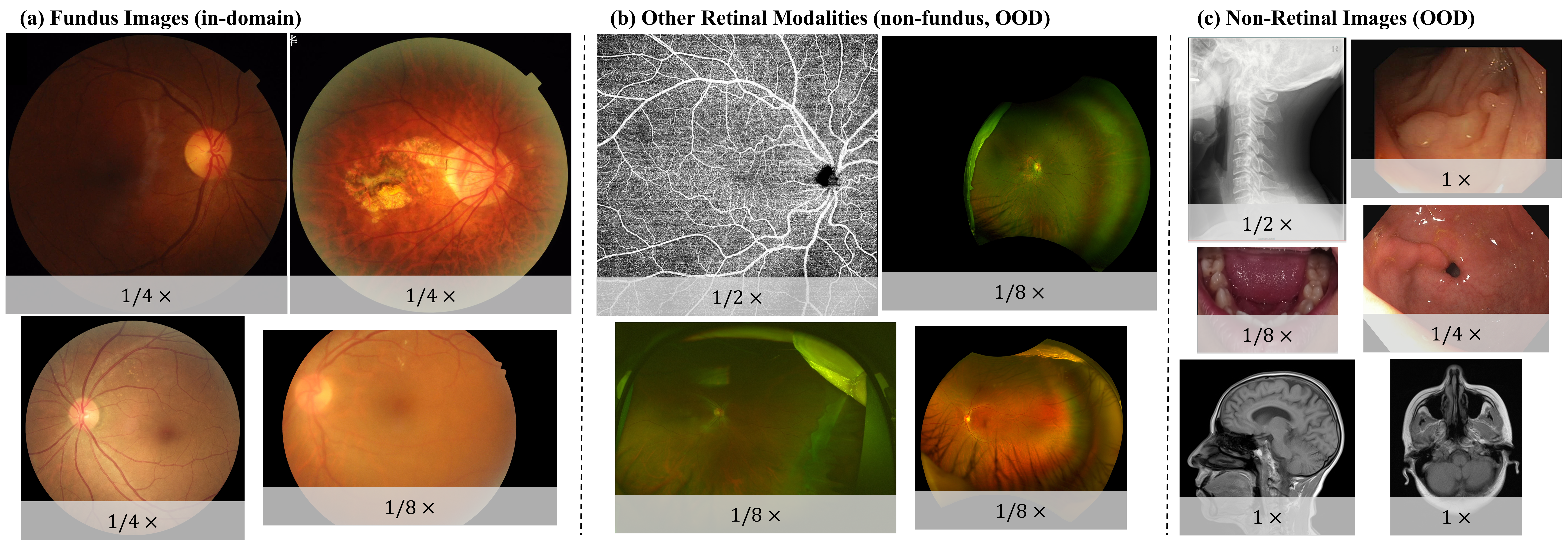}
  \caption{\textbf{Illustrative examples of imaging modalities considered for out-of-distribution (OOD) detection.}
  (a) Color fundus photography (CFP, in-domain) acquired using dedicated retinal cameras.
  (b) Other retinal imaging modalities, including fluorescein angiography and ultra-wide field imaging, which are treated as non-fundus images in this study.
  (c) Non-retinal medical images, representing OOD samples. 
  The overlaid scale factors (e.g., $1\times$, $1/2\times$, $1/4\times$, $1/8\times$) represent relative display scales corresponding to the actual image resolutions.}
  \label{fig:examples}
\end{figure*}

\section{Methods}
\label{sec:methods}

\subsection{Datasets}
The representative task in this study was to classify each image as either fundus or non-fundus. A total of 61{,}143 images from publicly available datasets were used for model training and evaluation, categorized into two groups: 
internal validation set (IV) and external validation set (EV). 
The IV set refers to all datasets used for model training and internal evaluation, while the EV set was strictly held out for test-only evaluation.

Seven publicly available color fundus photography (CFP) datasets, namely 
ADAM~\cite{dt4f-rt59-20}, FIVES~\cite{jin2022fives}, G1020~\cite{bajwa2020g1020},
MESSIDOR~\cite{ardakani2023open}, PAPILA~\cite{kovalyk2022papila}, 
REFUGE~\cite{tz6e-r977-19}, and RFMiD~\cite{data6020014}, were included in the IV set.  
In addition to these fundus datasets, the IV set also covered non-fundus and retinal OOD modalities such as 
fluorescein angiography (FA) and ultra-wide-field (UWF) fundus images from the DRAC dataset~\cite{qian2024drac,sun2025ultra}, 
brain MRI~\cite{brima2021brain}, cervical spine~\cite{ran2024high}, chest X-ray~\cite{nguyen2022vindr}, dental and oral images~\cite{chaudhary2024teeth}, 
skin lesion photographs~\cite{matin2023skin}, and endoscopic datasets such as CVC-Clinic~\cite{bernal2015wm} and Hyper-Kvasir~\cite{Borgli2020}.  

The EV set was used exclusively for independent testing to assess model generalization on unseen data.  
Specifically, the EV set included TOP~\cite{engelmann2022detection}, IDRiD~\cite{h25w98-18}, skin lesion photographs~\cite{matin2023skin}, and CVC-Clinic~\cite{bernal2015wm}, encompassing both fundus and non-fundus modalities.  
These datasets were not used during model training and served solely as test-only sources.

For model development and evaluation, the IV set, comprising a total of 45{,}909 images, was divided into training, validation, and testing subsets at a 3:1:1 ratio.
The EV set consisted of 15{,}234 images.
The IV set included 9{,}108 fundus and 36{,}801 non-fundus images, whereas the EV set contained 597 fundus and 14{,}637 non-fundus images.

\subsection{Preprocessing}
For ML models, original images were used without center cropping. Instead, images were analyzed at multiple relative resolutions ($1{\times}$, $1/2{\times}$, $1/4{\times}$, and $1/8{\times}$) to evaluate scale-dependent feature behavior. Grayscale ($Y{=}0.299R{+}0.587G{+}0.114B$) and HSV representations were generated when required for feature extraction. Local Binary Pattern (LBP) features were computed on $Y$ with $(P{=}8, R{=}1)$ using uniform mapping.
For DL models, all input images were center-cropped to the maximal square region and resized to multiple resolutions ($512{\times}512$, $256{\times}256$, $128{\times}128$, and $64{\times}64$) to assess resolution-dependent performance. RGB values were normalized to $[0,1]$, ensuring consistent input scaling for convolutional processing.

\begin{table*}[!bt]
\centering
\caption{
Comparison of end-to-end latency and accuracy across image resolutions for the ML and DL pipelines.
Latency is divided into Load (L), Feature extraction (F), and Classification (C) for ML, and into Load (L), Preprocessing (P), and Forward propagation (F).
Load (L) includes image and model loading.
All measurements were performed on 1{,}000 randomly sampled test images (batch size = 1) under single-image inference conditions.
Accuracy is reported for both internal (IV) and external (EV) validation sets.
Bold numbers denote the best-performing configuration and its associated latency for each pipeline (ML and DL).
}

\label{tab:result}
\setlength{\tabcolsep}{5.5pt}
\begin{tabular}{cccccccccccccccccc}
\toprule
\multicolumn{2}{c}{Resolution} &
\multicolumn{4}{c}{ML-CPU (ms)} &
\multicolumn{4}{c}{DL-CPU (ms)} &
\multicolumn{4}{c}{DL-GPU (ms)} &
\multicolumn{2}{c}{ML (accuracy)} &
\multicolumn{2}{c}{DL (accuracy)} \\
\cmidrule(lr){1-2}\cmidrule(lr){3-6}\cmidrule(lr){7-10}\cmidrule(lr){11-14}\cmidrule(lr){15-16}\cmidrule(lr){17-18}
ML & DL & L & F & C & Overall & L & P & F & Overall & L & P & F & Overall & IV & EV & IV & EV \\
\midrule
1$\times$   & $512{\times}512$ & 15 & 378 & 3 & 396 & 177 & 8 & 44 & \textbf{229} & 198 & 9 & 7 & \textbf{214} & 1.000 & 0.988 & \textbf{1.000} & \textbf{1.000} \\
1/2$\times$ & $256{\times}256$ & 15 & 101 & 3 & 119 & 175 & 6 & 16 & \textbf{197} & 187 & 5 & 4 & \textbf{196} & 1.000 & 0.985 & \textbf{1.000} & \textbf{1.000} \\
1/4$\times$ & $128{\times}128$ & 15 & 32  & 3 & 50  & 173 & 5 & 10 & 188 & 188 & 4 & 4 & 196 & 1.000 & 0.995 & 1.000 & 0.995 \\
1/8$\times$ & $64{\times}64$   & 15 & 13  & 3 & \textbf{31}  & 173 & 4 & 6 & 183 & 191 & 3 & 3 & 197 & \textbf{1.000} & \textbf{0.999} & 1.000 & 0.997 \\
\bottomrule
\end{tabular}
\end{table*}

\subsection{Machine Learning Pipeline and Hand-crafted Features}

We extracted 39 hand-crafted features, organized into five categories: intensity and background statistics, color and texture, spatial distribution, shape and morphology, and global structural indicators. 
These features were designed to capture domain-specific characteristics of CFP, including its circular field of view, dark peripheral background, and reddish–orange hue distribution. 
They provide transparent and clinically interpretable cues for distinguishing fundus images from non-fundus images, including both retinal and non-retinal modalities.

The feature categories are summarized as follows:  
(1) \textbf{Intensity and background statistics}: measure global brightness, contrast, and peripheral darkness (e.g., \textit{min}, \textit{max}, \textit{mean}, \textit{variance}, \textit{background flag});  
(2) \textbf{Color and texture}: capture chromatic consistency and fine-grained texture using hue/saturation statistics and local descriptors such as LBP and GLCM (e.g., \textit{hue mean}, \textit{sat var}, \textit{glcm contrast}, \textit{lbp mean});  
(3) \textbf{Spatial distribution}: represent center–periphery intensity gradients and hemispheric brightness asymmetries (e.g., \textit{radial slope mean}, \textit{center minus outer}, \textit{hemisphere asymmetry});  
(4) \textbf{Shape and morphology}: capture geometric characteristics of the retinal mask that reflect the circular shape of fundus images, including measures of circularity, eccentricity, solidity, and boundary irregularity (e.g., \textit{circularity}, \textit{solidity}, \textit{circle residual mean}, \textit{eccentricity}). 
Most of these features are designed to quantify how closely the observed retinal region follows the typical round structure of CFP, thereby serving as strong indicators for distinguishing in-domain fundus samples from OOD or non-retinal images; and  
(5) \textbf{Global structural indicators}: summarize overall image composition, including color type and the proportion of dark regions (e.g., \textit{is rgb}, \textit{black pixel ratio}).

An Extremely Randomized Trees (\textit{ExtraTrees}) \cite{geurts2006extremely} classifier was trained using 100 estimators, a maximum depth of 10, and the square-root feature subset strategy. 
To balance generalization and model simplicity, each internal split required at least 10 samples, and leaf nodes contained a minimum of two samples. 
Inference was executed on a single CPU thread to enable accurate measurement of per-image cold-start latency. 
The model was implemented using the \textit{scikit-learn} library. 
For interpretability, SHapley Additive exPlanations (SHAP) \cite{lundberg2017unified} values were computed to quantify feature-level contributions for each image.

\subsection{Deep learning pipeline}

A ResNet-18 backbone pretrained on ImageNet was employed as the DL baseline. 
The network was fine-tuned using the AdamW optimizer with a learning rate of $1\times10^{-4}$, 
weight decay of $1\times10^{-4}$, a batch size of 128, and trained for 3 epochs. 
The best checkpoint was selected according to the minimum validation loss, and cross-entropy loss 
was used as the optimization objective.

All images were center-cropped according to the shorter side of width or height, 
then resized to the target training resolution. 
Each image was standardized using the ImageNet mean and standard deviation.
To improve model robustness and generalization, probabilistic data augmentations were applied, 
including horizontal ($p=0.5$) and vertical ($p=0.2$) flips, 
rotations up to $\pm15^{\circ}$, and photometric variations such as 
brightness/contrast adjustments and gamma correction within $\pm20\%$.

Grad-CAM++~\cite{chattopadhay2018grad} was employed to visualize the discriminative image regions 
that contributed most to the model’s decision. 
The activation maps were extracted from the final convolutional block of the ResNet-18 backbone, and computed with respect to each image’s ground-truth label. 
The resulting heatmaps were blended with the corresponding input images 
to qualitatively assess the spatial correspondence between model attention and clinically relevant areas.

\subsection{Evaluation metrics and latency measurement}
A five-fold cross-validation was conducted for internal validation using stratified sampling to preserve class balance across folds. 
In each fold, the model was trained on four subsets and evaluated on the remaining test subset. 
After fold-wise testing, predictions from all five test subsets were concatenated to compute the overall internal validation performance. 
For external validation, the five models obtained from cross-validation were independently applied to an unseen external dataset, and the predicted probabilities for each image were averaged across models to obtain a single consensus prediction. 
A fixed decision threshold of $\tau = 0.5$ was applied uniformly across all experiments to produce binary predictions. 
Model performance was evaluated using the area under the receiver operating characteristic curve (AUROC) and accuracy.

To assess computational efficiency, inference latency was systematically measured for both ML and DL pipelines. 
For the ML pipeline, latency was decomposed into model loading, feature extraction, and classification. 
For the DL baseline, latency was measured separately on CPU and GPU, consisting of model loading, preprocessing, and forward propagation through the encoder. 
End-to-end latency was defined as the total elapsed time from model loading to the generation of the final probability output. 
Latency measurements were performed on 1{,}000 randomly sampled test images with a batch size of one. 
The ML pipeline was tested at multiple downsampling ratios ($\times1$, $\times1/2$, $\times1/4$, and $\times1/8$), while the DL model was evaluated at the corresponding input resolutions ($512^2$, $256^2$, $128^2$, and $64^2$).
All experiments were executed on a workstation equipped with two Intel Xeon Gold 5218R CPUs (80 cores, 2.10 GHz), 504 GB RAM, and an NVIDIA Tesla V100 GPU (32 GB VRAM).

\section{Results}
\label{sec:results}

\subsection{Cross-validation and external validation performance}
Table~\ref{tab:result} summarizes the quantitative results across image resolutions.
Both the ML (ExtraTrees) and DL (ResNet-18) models achieved AUROC values of 1.000 under all conditions.
Accuracy differences between the two approaches were marginal across internal (IV) and external (EV) validation sets.
For the ML pipeline, accuracy ranged from 0.985 to 1.000 across resolutions, whereas the DL model ranged from 0.995 to 1.000.
At the lowest resolution ($64{\times}64$, $\times1/8$), the ML model achieved 1.000 on the internal set and 0.999 on the external set, effectively matching the DL baseline despite its lightweight design.
Interestingly, the ML pipeline maintained robust performance even at lower resolutions, whereas the DL model showed a decrease in accuracy as image resolution decreased.

\subsection{Resolution and latency analysis}
As shown in Table~\ref{tab:result}, the ML and DL pipelines exhibited distinct scaling behaviors with respect to image resolution.
For the ML model, latency at the highest configuration ($\times1$) reached 396~ms on CPU, primarily due to feature extraction (378~ms). 
As resolution decreased, ML latency dropped sharply—from 396~ms at $\times1$ to 31~ms at $\times1/8$—while maintaining comparable accuracy (0.999 at EV). 
In contrast, the DL model showed consistently high latency across resolutions (183–229~ms) owing to its fixed architectural complexity and large parameter count, with runtime dominated by model loading, memory allocation, and I/O overhead rather than input-dependent computation.

These results demonstrate that the ML pipeline scales efficiently with image size and achieves substantial speed gains without sacrificing accuracy, exhibiting virtually no accuracy–latency trade-off. 
By comparison, the DL model’s latency remains largely invariant to resolution due to fixed initialization and memory transfer costs. 
Overall, the ML pipeline provides equivalent discrimination with up to a sixfold latency reduction, highlighting its suitability for lightweight, real-time OOD detection in clinical settings with limited computational resources.

\FloatBarrier 
\begin{figure}[!htbp]
  \centering
  \includegraphics[width=\columnwidth]{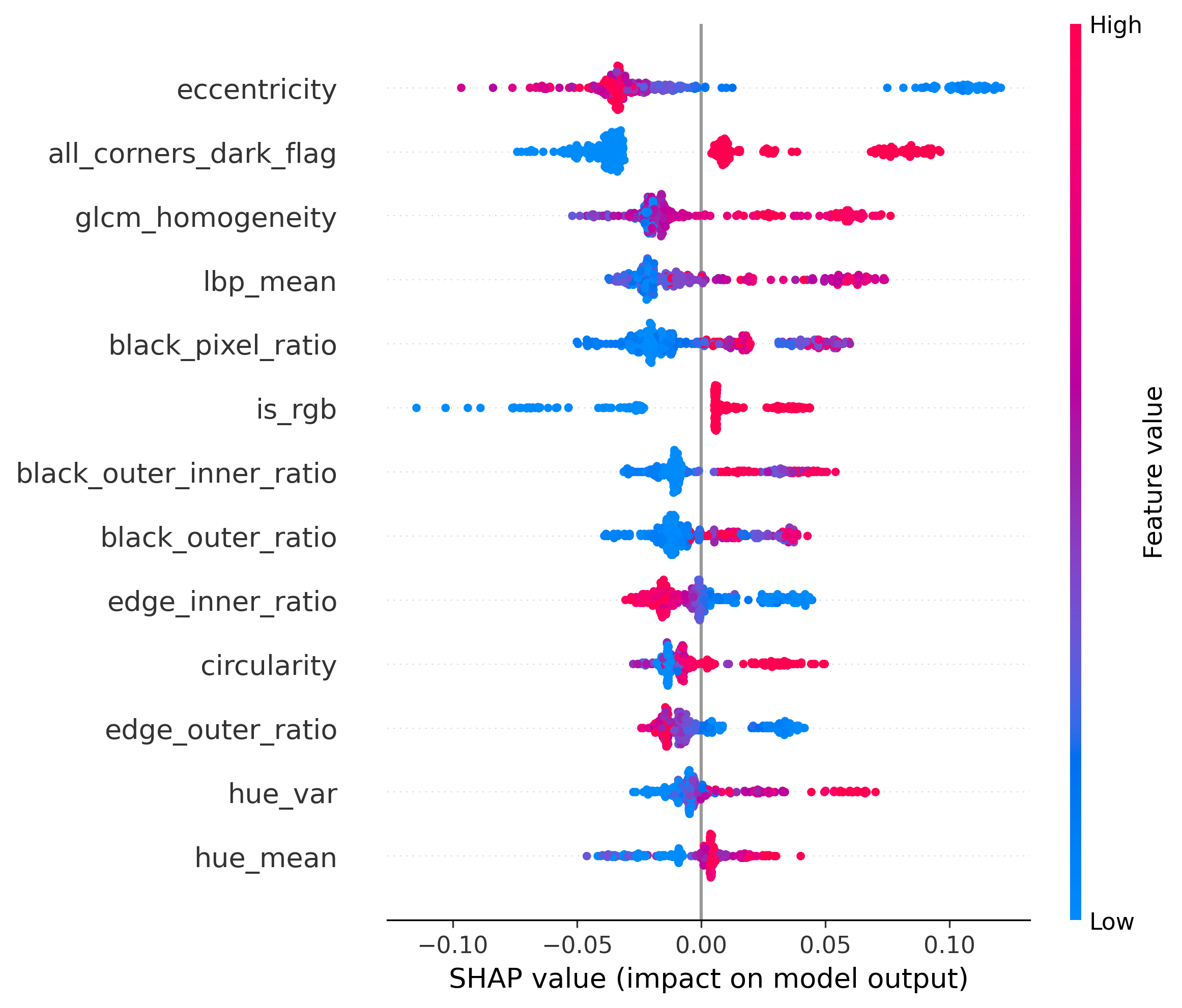}
  \caption{\textbf{SHAP summary plot.}
  SHAP value distributions showing the contribution of each feature to model output.}
  \label{fig:shap}
\end{figure}

\begin{figure}[t]
  \centering
  \includegraphics[width=\columnwidth]{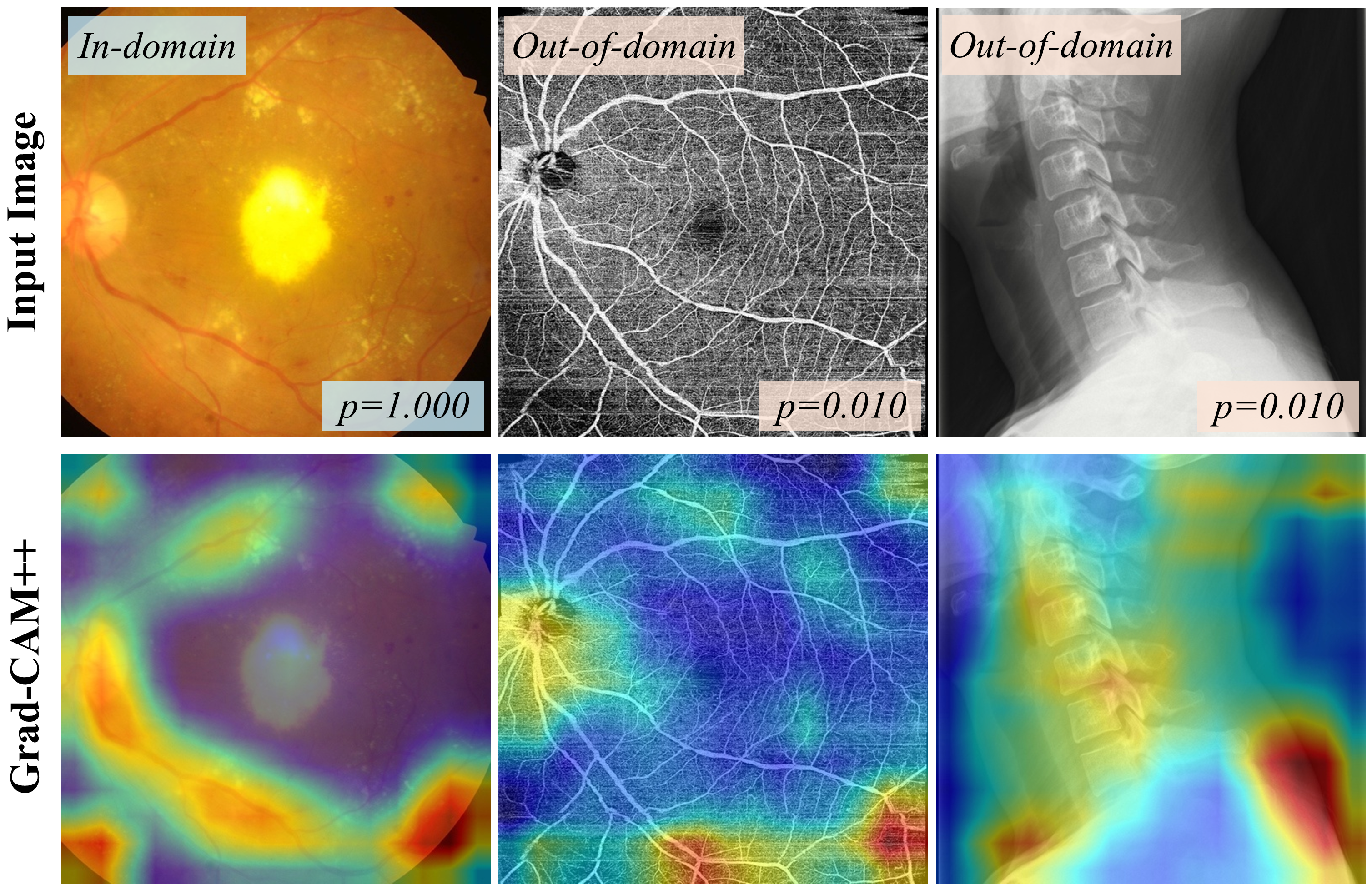}
\caption{\textbf{Grad-CAM++ visualizations (ResNet-18).}
The first two images, both retinal in origin, show similar activations around the optic disc and vessels despite belonging to different domains, indicating that these responses are unrelated to OOD discrimination. 
For the X-ray image, activations are spatially diffuse and lack any meaningful correspondence.}
  \label{fig:cam}
\end{figure}

\subsection{Interpretability: SHAP for ML vs.\ CAM for DL}
The SHAP summary plot (Figure~\ref{fig:shap}) reveals that \textit{eccentricity}, \textit{all\_corners\_dark\_flag}, and \textit{glcm\_homogeneity} are the three most influential features in distinguishing fundus from non-fundus images. A low \textit{eccentricity} value, corresponding to a near-circular shape, contributes strongly to the “fundus” prediction, consistent with the circular geometry characteristic of retinal images. Similarly, a high \textit{all\_corners\_dark\_flag} value positively impacts the “fundus” class, reflecting the presence of a dark peripheral border typical of fundus photographs. In addition, high \textit{glcm\_homogeneity} values are associated with smooth and uniform textures, further reinforcing the “fundus” classification. Together, these findings indicate that the ML model primarily relies on interpretable global cues—circularity, peripheral shading, and texture uniformity—to identify fundus images.

While the ML pipeline enables feature-level interpretation through SHAP, the DL model depends on gradient-based saliency visualization such as Grad-CAM++. 
However, in binary OOD detection, these visualizations offer little insight into the actual decision process. Figure~\ref{fig:cam} shows Grad-CAM++ results from the ResNet-18 baseline for representative in-domain and OOD samples. 
The first two images are both retinal in origin—fundus and fluorescein angiography—yet the activation maps exhibit nearly identical patterns, concentrating around the optic disc and major vessels. 
Such regions are visually distinctive but unrelated to the OOD classification itself, indicating that the model attends to generic high-contrast structures rather than domain-specific cues. 
For the non-retinal X-ray image, activations are spatially diffuse and lack consistent anatomical meaning. 
Overall, gradient-based explainability methods that are effective in lesion or anomaly localization tasks provide limited interpretive value for OOD detection.

\section{Discussion}
\label{sec:disc}

Out-of-distribution (OOD) detection is a critical prerequisite for the reliable deployment of AI systems in clinical practice.
In retinal imaging, verifying input validity before diagnostic inference is essential, as predictions on non-fundus images can degrade accuracy and compromise user trust. Effective OOD filtering (i.e., fundus vs. non-fundus screening) ensures that downstream diagnostic models operate only on valid data, thereby improving safety, traceability, and overall system integrity.

The experimental results indicate that, under the studied setting, distinguishing fundus from non-fundus images forms a well-defined classification problem when trained on sufficiently diverse datasets.
Both the ML and DL models achieved near-saturated performance (AUROC = 1.000, ACC $\geq$ 0.999) on both internal and external validation sets. 
Under such conditions, additional architectural complexity provides no measurable performance gain in this task setting, and practical deployment factors—computation time, memory demand, and interpretability—become the dominant criteria for model selection.

The proposed \emph{ExtraTrees}-based ML pipeline, built on a compact set of hand-crafted and fundus-aware descriptors, achieved sub-\SI{31}{ms} end-to-end latency on CPU-only hardware while maintaining detection accuracy comparable to the DL baseline. 
SHAP analysis revealed stable and domain-consistent feature attributions—highlighting the model's reliance on peripheral darkness, smooth color distribution, and circular image geometry—thereby enabling transparent and reproducible interpretation of its decision process. 
These characteristics make the ML pipeline particularly suitable for real-time OOD filtering in clinical systems, where deterministic behavior and limited computational resources are critical constraints.

In contrast, the \emph{ResNet-18} baseline required substantially greater computational and memory resources. 
At the lowest resolution ($64{\times}64$), its end-to-end inference latency was approximately 183~ms on CPU and 197~ms on GPU, compared to 31~ms for the ML pipeline under identical conditions—a roughly sixfold increase. 
For example, commercial software such as \emph{VUNO Med--Fundus AI}, which performs full lesion detection and quantitative grading across twelve categories, operates in under 2 seconds on CPU-only hardware. 
Given that OOD detection involves only binary filtering rather than full diagnostic inference, a latency of nearly 200~ms per image represents a significant overhead. 
Moreover, all latency measurements in this study were obtained on a high-spec workstation. In typical clinical environments using standard desktop hardware, the runtime gap between ML and DL approaches would likely be even more pronounced.
Despite the higher computational cost, the DL model offered no measurable improvement in accuracy or AUROC relative to the ML approach across all evaluation settings. 
These findings demonstrate that the compact ML model achieves equivalent discrimination capability while maintaining substantially higher computational efficiency and predictable latency.

This study has several limitations.
First, although the datasets cover diverse imaging modalities, they may not capture rare acquisition scenarios such as handheld or smartphone-based fundus imaging, or extreme peripheral distortion.
Second, the reported latency values may vary across systems depending on hardware and software configurations, and should therefore be interpreted as indicative rather than absolute.
Finally, the current feature set is optimized for fundus imagery and would require domain-specific adaptation for other medical imaging applications.

In conclusion, a lightweight and interpretable ML model can deliver OOD detection performance equivalent to a fine-tuned DL model while achieving far greater computational efficiency. These results support a practical deployment architecture in which ML-based OOD filtering operates at the data-ingestion stage, ensuring input validity and reserving DL computation for downstream diagnostic analysis.

\section{COMPLIANCE WITH ETHICAL STANDARDS}
This research study was conducted retrospectively using human subject data made available in open access. Ethical approval was not required as confirmed by the license attached with the open access data.

\section{Conflict of Interest}
The authors are employees of VUNO Inc., but declare that they have no competing financial or non-financial interests related to this work.

\bibliographystyle{IEEEtran}
\bibliography{refs}
\end{document}